%% file: main.tex
\documentclass{article}

\usepackage{microtype}
\usepackage{graphicx}
\usepackage{subfigure}
\usepackage{booktabs} %

\usepackage[backref=page]{hyperref}

\usepackage[accepted]{icml2025}

\usepackage{amsmath}
\usepackage{amssymb}
\usepackage{mathtools}
\usepackage{amsthm}
\usepackage{thmtools, thm-restate}
\usepackage{enumerate,enumitem}
\usepackage{tcolorbox}
\usepackage{multirow}

\theoremstyle{plain}

\theoremstyle{definition}
\newtheorem{definition}{Definition}[section]

\newtheorem{researchquestion}{Question}[subsection]
\theoremstyle{remark}

\usepackage[textsize=tiny]{todonotes}

\usepackage{url}

\usepackage{breakurl}

\usepackage{makecell}
\usepackage[capitalize,noabbrev,nameinlink]{cleveref}
\usepackage{crossreftools}

\crefname{section}{\S}{\S}
\crefname{subsection}{\S}{\S}
\crefname{subsubsection}{\S}{\S}
\crefname{figure}{Fig.}{Figs.}
\crefname{prop}{Prop.}{Props.}
\crefname{proposition}{Prop.}{Props.}
\crefname{appendix}{Appx.}{Appxs.}
\crefname{algorithm}{Alg.}{Algs.}
\crefname{theorem}{Thm.}{Thms.}
\crefname{conjecture}{Conj.}{Conjs.}
\crefname{researchquestion}{Q.}{Qs.}
\crefname{definition}{Defn.}{Defns.}
\crefname{cor}{Cor.}{Cors.}
\crefname{lem}{Lem.}{Lems.}
\crefname{table}{Tab.}{Tabs.}
\crefname{assum}{Assum.}{Assums.}
\crefname{example}{Ex.}{Exs.}

\input{commands}
\input{glossary}
\input{acronyms}

\usepackage{pifont}
\usepackage{svg}

\definecolor{figblue}{HTML}{4A90E2}
\definecolor{figred}{HTML}{D0021B}
\definecolor{figgreen}{HTML}{2CA02C}

\everypar{\looseness=-1}

\begin{document}

\twocolumn[
\icmltitle{Position: 
An Empirically Grounded Identifiability Theory Will Accelerate Self-Supervised Learning Research
}

\icmltitlerunning{Position: 
An Empirically Grounded Identifiability Theory Will Accelerate Self-Supervised Learning Research
}

\icmlsetsymbol{equal}{*}

\begin{icmlauthorlist}
\icmlauthor{Patrik Reizinger}{tueb,ellis}
\icmlauthor{Randall Balestriero}{brown}
\icmlauthor{David Klindt}{cshl}
\icmlauthor{Wieland Brendel}{tueb,ellis,tuai}

\end{icmlauthorlist}

\icmlaffiliation{tueb}{Max Planck Institute for Intelligent Systems, Tübingen, Germany}
\icmlaffiliation{ellis}{ELLIS Institute Tübingen, Germany}
\icmlaffiliation{tuai}{Tübingen AI Center, Germany}
\icmlaffiliation{brown}{Department of Computer Science, Brown University, Rhode Island, USA}
\icmlaffiliation{cshl}{Cold Spring Harbor Laboratory, Cold Spring Harbor, New York, USA}

\icmlcorrespondingauthor{Patrik Reizinger}{patrik.reizinger@tuebingen.mpg.de}

\icmlkeywords{Machine Learning, ICML}

\vskip 0.3in
]

\printAffiliationsAndNotice{}  %

\begin{abstract}

\input{abstract.tex}
\end{abstract}

\input{main_text.tex}
\section*{Acknowledgements}

The authors would like to thank Laure Ciernik, Evgenia Rusak, Roland Zimmermann, Julius von Kügelgen, Michael Kirchhof, Lucas Maes, Vishaal Udandarao, Ameya Prabhu, Karsten Roth, Omar Chehab, Runtian Zhai, Kamilė Stankevičiūtė, Thaddäus Wiedemer, Pradeep Ravikumar, Luigi Gresele, Quentin Garrido, Alice Bizeul, Mark Ibrahim, and Attila Juhos for their valuable suggestions for shaping the manuscript.
The authors thank the International Max Planck
Research School for Intelligent Systems (IMPRS-IS) for supporting Patrik Reizinger. Patrik Reizinger acknowledges his membership in the European Laboratory for Learning and Intelligent Systems (ELLIS) PhD program. This work was supported by the German Federal Ministry of Education and Research (BMBF): Tübingen AI Center, FKZ: 01IS18039A. Wieland Brendel acknowledges financial support via an Emmy Noether Grant funded by the German Research Foundation (DFG) under grant no. BR 6382/1-1 and via the Open Philantropy Foundation funded by the Good Ventures Foundation. Wieland Brendel is a member of the Machine Learning Cluster of Excellence, EXC number 2064/1 – Project number 390727645. This research utilized compute resources at the Tübingen Machine Learning Cloud, DFG FKZ INST 37/1057-1 FUGG.

\section*{Impact Statement}

This paper presents work whose goal is to advance the field of 
Machine Learning. There are many potential societal consequences 
of our work, none which we feel must be specifically highlighted here.

\bibliographystyle{icml2025}
\bibliography{references,references2}

\newpage
\appendix
\onecolumn 
\input{appendix}
\printglossaries

\end{document}

%% file: commands.tex
\newcommand{\ie}{i.e.\@\xspace}
\newcommand{\Ie}{I.e.\@\xspace}
\newcommand{\eg}{e.g.\@\xspace}

\newcommand{\etc}{etc.\@\xspace}

\newcommand{\cf}{cf.\@\xspace} %

\newcommand{\wrt}{w.r.t.\@\xspace} %

\newcommand{\xmark}{\ding{55}}%

\newcommand{\inv}[1]{\ensuremath{#1^{-1}}}

\newcommand{\mat}[1]{\ensuremath{\boldsymbol{\mathrm{#1}}}}

\newcommand{\abs}[1]{\ensuremath{\left|#1\right|}}

\newcommand{\expectation}[1]{\ensuremath{\mathbb{E}_{#1}}}
\newcommand{\rr}[1]{\ensuremath{\mathbb{R}^{#1}}}

\newcommand{\parenthesis}[1]{\ensuremath{\left(#1\right)}}

 \renewcommand{\algorithmiccomment}[1]{//#1}

%% file: glossary.tex
\usepackage[acronym, automake, toc, nomain, nopostdot, style=tree, nonumberlist,numberedsection]{glossaries}
\usepackage{glossary-mcols}
\usepackage{xparse}
\usepackage{xspace}
\setglossarystyle{mcolindex}

\newglossary{abbrev}{abs}{abo}{Nomenclature}

\newglossaryentry{aux}{
    name        = \ensuremath{\mathrm{\boldsymbol{u}}} ,
    description = {auxiliary variable} ,
    type        = abbrev,
}

\newglossaryentry{im}{
    name        = \ensuremath{\mathrm{Im}} ,
    description = {image space} ,
    type        = abbrev,
}

\newglossaryentry{ker}{
    name        = \ensuremath{\mathrm{Ker}} ,
    description = {kernel space} ,
    type        = abbrev,
}

\newglossaryentry{r2}{
    name        = \ensuremath{R^2} ,
    description = {coefficient of determination} ,
    type        = abbrev,
}

\newglossaryentry{kronecker}{
    name        = \ensuremath{\otimes} ,
    description = {Kronecker product} ,
    type        = abbrev,
}

\newglossaryentry{loss}{
    name        = \ensuremath{\mathcal{L}} ,
    description = {loss function} ,
    type        = abbrev,
}

\newglossaryentry{numenv}{
    name        = \ensuremath{\abs{E}} ,
    description = {number of environments} ,
    type        = abbrev,
}

\newglossaryentry{lr}{
    name        = \ensuremath{\eta} ,
    description = {learning rate} ,
    type        = abbrev,
}

\newglossaryentry{hypersphere}{
    name        = \ensuremath{\mathcal{S}} ,
    description = {hypersphere} ,
    type        = abbrev,
}

\newglossaryentry{dec}{
    name        = \ensuremath{\boldsymbol{f}} ,
    description = {decoder map $\gls{Latent}\to\gls{Obs}$} ,
    type        = abbrev,
}

\newglossaryentry{deccomp}{
    name        = \ensuremath{f} ,
    description = {decoder map component} ,
    type        = abbrev,
}

\newglossaryentry{enc}{
    name        = \ensuremath{\boldsymbol{g}} ,
    description = {encoder map $\gls{Obs}\to\gls{Latent}$} ,
    type        = abbrev,
}

\newglossaryentry{numdata}{
    name        = \ensuremath{n} ,
    description = {number of samples} ,
    type        = abbrev,
}

\newglossaryentry{observations}{type=abbrev,name=Observations,description={\nopostdesc}}

\newglossaryentry{obs}{
    name        = \ensuremath{\boldsymbol{x}} ,
    description = {observation vector} ,
    type        = abbrev,
    parent      = observations,
}

\newglossaryentry{obscomp}{
    name        = \ensuremath{x} ,
    description = {observation single component} ,
    type        = abbrev,
    parent      = observations,
}

\newglossaryentry{Obs}{
    name        = \ensuremath{\mathcal{X}} ,
    description = {observation space} ,
    type        = abbrev,
    parent      = observations,
}

\newglossaryentry{obsdim}{
    name        = \ensuremath{D} ,
    description = {dimensionality of the observation space \gls{Obs}} ,
    type        = abbrev,
    parent      = observations,
}

\newglossaryentry{obsmat}{
    name        = \ensuremath{\mat{X}} ,
    description = {observation matrix of \rr{\gls{numdata}\times\gls{obsdim}}} ,
    type        = abbrev,
    parent      = observations,
}

\newglossaryentry{obspos}{
    name        = \ensuremath{\tilde{\boldsymbol{x}}} ,
    description = {positive observation vector} ,
    type        = abbrev,
    parent      = observations,
}

\newglossaryentry{obsneg}{
    name        = \ensuremath{{\boldsymbol{x}}^{-}} ,
    description = {negative observation vector} ,
    type        = abbrev,
    parent      = observations,
}

\newglossaryentry{labels}{type=abbrev,name=Labels,description={\nopostdesc}}

\newglossaryentry{label}{
    name        = \ensuremath{\boldsymbol{y}} ,
    description = {label vector} ,
    type        = abbrev,
    parent      = labels,
}

\newglossaryentry{labelhat}{
    name        = \ensuremath{\widehat{\boldsymbol{y}}} ,
    description = {estimated label vector} ,
    type        = abbrev,
    parent      = labels,
}

\newglossaryentry{labelcomp}{
    name        = \ensuremath{y} ,
    description = {label component} ,
    type        = abbrev,
    parent      = labels,
}

\newglossaryentry{labelcomphat}{
    name        = \ensuremath{\widehat{y}} ,
    description = {label component} ,
    type        = abbrev,
    parent      = labels,
}

\newglossaryentry{labelset}{
    name        = \ensuremath{\mathcal{Y}} ,
    description = {label set} ,
    type        = abbrev,
    parent      = labels,
}

\newglossaryentry{labeldim}{
    name        = \ensuremath{C} ,
    description = {number of classes in the label set \gls{labelset}} ,
    type        = abbrev,
    parent      = labels,
}

\newglossaryentry{latents}{type=abbrev,name=Latents,description={\nopostdesc}}

\newglossaryentry{latent}{
    name        = \ensuremath{\boldsymbol{z}} ,
    description = {latent vector} ,
    type        = abbrev,
    parent     = latents,
}

\newglossaryentry{latentcomp}{
    name        = \ensuremath{z} ,
    description = {latent single component} ,
    type        = abbrev,
    parent     = latents,
}

\newglossaryentry{Latent}{
    name        = \ensuremath{\mathcal{Z}} ,
    description = {latents} ,
    type        = abbrev,
    parent     = latents,
}

\newglossaryentry{latentdim}{
    name        = \ensuremath{d} ,
    description = {dimensionality of the latent space \gls{Latent}} ,
    type        = abbrev,
    parent     = latents,
}

\newglossaryentry{latentmat}{
    name        = \ensuremath{\mat{Z}} ,
    description = {latent matrix of \rr{\gls{numdata}\times\gls{latentdim}}} ,
    type        = abbrev,
    parent      = latents,
}

\newglossaryentry{latentpos}{
    name        = \ensuremath{\tilde{\boldsymbol{z}}} ,
    description = {positive latent vector} ,
    type        = abbrev,
    parent      = latents,
}

\newglossaryentry{latentneg}{
    name        = \ensuremath{\boldsymbol{z}^{-}} ,
    description = {negative latent vector} ,
    type        = abbrev,
    parent      = observations,
}

\newglossaryentry{sigmaz}{
    name        = \ensuremath{\boldsymbol{\sigma}_{\gls{latentcomp}}} ,
    description = {std of \gls{latentcomp}} ,
    type        = abbrev,
    parent     = latents,
}

\newglossaryentry{content}{
    name        = \ensuremath{\boldsymbol{z}^{c}} ,
    description = {content latent vector} ,
    type        = abbrev,
    parent     = latents,
}

\newglossaryentry{contentcomp}{
    name        = \ensuremath{z^{c}} ,
    description = {content latent single component} ,
    type        = abbrev,
    parent     = latents,
}

\newglossaryentry{Content}{
    name        = \ensuremath{\mathcal{Z}^{c}} ,
    description = {content} ,
    type        = abbrev,
    parent     = latents,
}

\newglossaryentry{contentdim}{
    name        = \ensuremath{d_{c}} ,
    description = {dimensionality of \gls{content}} ,
    type        = abbrev,
    parent     = latents,
}

\newglossaryentry{sigmac}{
    name        = \ensuremath{\boldsymbol{\sigma}_{c}} ,
    description = {std of \gls{contentcomp}} ,
    type        = abbrev,
    parent     = latents,
}

\newglossaryentry{style}{
    name        = \ensuremath{\boldsymbol{z}^{s}} ,
    description = {style latent vector} ,
    type        = abbrev,
    parent     = latents,
}

\newglossaryentry{stylecomp}{
    name        = \ensuremath{z^{s}} ,
    description = {style latent single component} ,
    type        = abbrev,
    parent     = latents,
}

\newglossaryentry{Style}{
    name        = \ensuremath{\mathcal{Z}^{s}} ,
    description = {style} ,
    type        = abbrev,
    parent     = latents,
}

\newglossaryentry{styledim}{
    name        = \ensuremath{d_{s}} ,
    description = {dimensionality of \gls{style}} ,
    type        = abbrev,
    parent     = latents,
}

\newglossaryentry{sigmas}{
    name        = \ensuremath{\boldsymbol{\sigma}_{s}} ,
    description = {std of \gls{stylecomp}} ,
    type        = abbrev,
    parent     = latents,
}

\newglossaryentry{modality}{
    name        = \ensuremath{\boldsymbol{z}^{m}} ,
    description = {modality-specific latent vector} ,
    type        = abbrev,
    parent     = latents,
}

\newglossaryentry{modalitycomp}{
    name        = \ensuremath{z^{m}} ,
    description = {modality-specific  latent single component} ,
    type        = abbrev,
    parent     = latents,
}

\newglossaryentry{Modality}{
    name        = \ensuremath{\mathcal{Z}^{m}} ,
    description = {latent subspace of \gls{modality}} ,
    type        = abbrev,
    parent     = latents,
}

\newglossaryentry{modalitydim}{
    name        = \ensuremath{d_{m}} ,
    description = {dimensionality of \gls{modality}} ,
    type        = abbrev,
    parent     = latents,
}

\newglossaryentry{algebra}{type=abbrev,name=Algebra,description={\nopostdesc}}

\newglossaryentry{identity}{
    name        = \ensuremath{\boldsymbol{\mathrm{I}}} ,
    description = { identity matrix} ,
    type        = abbrev,
    parent      = algebra,
}

\newglossaryentry{ones}{
    name        = \ensuremath{\boldsymbol{\mathrm{1}}} ,
    description = {a vector of ones} ,
    type        = abbrev,
    parent      = algebra,
}

\newglossaryentry{zeros}{
    name        = \ensuremath{\boldsymbol{\mathrm{0}}} ,
    description = {a vector of zeros} ,
    type        = abbrev,
    parent      = algebra,
}

\newglossaryentry{jacobian}{
    name        = \ensuremath{\boldsymbol{\mathrm{J}}} ,
    description = {Jacobian matrix} ,
    type        = abbrev,
    parent      = algebra,
}

\newglossaryentry{hessian}{
    name        = \ensuremath{\boldsymbol{\mathrm{H}}} ,
    description = {Hessian matrix} ,
    type        = abbrev,
    parent      = algebra,
}

\newglossaryentry{d}{
    name        = \ensuremath{\boldsymbol{\mathrm{D}}} ,
    description = {diagonal matrix} ,
    type        = abbrev,
    parent      = algebra,
}

\newglossaryentry{o}{
    name        = \ensuremath{\boldsymbol{\mathrm{O}}},
    description = {orthogonal matrix} ,
    type        = abbrev,
    parent      = algebra,
}

\newglossaryentry{scalar}{
    name        = \ensuremath{\alpha} ,
    description = {scalar field} ,
    type        = abbrev,
    parent      = algebra,
}

\newglossaryentry{perm}{
    name        = \ensuremath{\mathbb{P}} ,
    description = {group of permutation matrices} ,
    type        = abbrev,
    parent      = algebra,
}

\newglossaryentry{p}{
    name        = \ensuremath{\mat{P}},
    description = {permutation matrix} ,
    type        = abbrev,
    parent      = algebra,
}

\newglossaryentry{prob}{type=abbrev,name=Probability theory,description={\nopostdesc}}

\newglossaryentry{cov}{
    name        = \ensuremath{\boldsymbol{\mathrm{\Sigma}}},
    description = {covariance matrix} ,
    type        = abbrev,
    parent      = prob,
}

\newglossaryentry{mean}{
    name        = \ensuremath{\boldsymbol{\mu}},
    description = {mean} ,
    type        = abbrev,
    parent      = prob,
}

\newglossaryentry{std}{
    name        = \ensuremath{\boldsymbol{\sigma}},
    description = {standard deviation} ,
    type        = abbrev,
    parent      = prob,
}

\newglossaryentry{entropy}{
    name        = \ensuremath{\mathrm{H}} ,
    description = {entropy} ,
    type        = abbrev,
    parent      = prob,
}

\newglossaryentry{expfamparam}{
    name        = \ensuremath{\boldsymbol{\theta}} ,
    description = {parameter of exponential family} ,
    type        = abbrev,
    parent      = prob,
}

\newglossaryentry{expfamnatparam}{
    name        = \ensuremath{\boldsymbol{\eta}} ,
    description = {natural parameter of exponential family} ,
    type        = abbrev,
    parent      = prob,
}

\newglossaryentry{expfamsuffstat}{
    name        = \ensuremath{T(\gls{obs})} ,
    description = {sufficient statistics of exponential family} ,
    type        = abbrev,
    parent      = prob,
}

\newglossaryentry{expfamlogpartition}{
    name        = \ensuremath{A} ,
    description = {log parition function of exponential family (depends on \gls{expfamnatparam})} ,
    type        = abbrev,
    parent      = prob,
}

\newglossaryentry{wishart}{
    name        = \ensuremath{\mathcal{W}} ,
    description = {Wishart distribution} ,
    type        = abbrev,
    parent      = prob,
}

\newglossaryentry{normal}{
    name        = \ensuremath{\mathcal{N}} ,
    description = {normal distribution} ,
    type        = abbrev,
    parent      = prob,
}

\newglossaryentry{matrixnormal}{
    name        = \ensuremath{\mathcal{MN}} ,
    description = {normal distribution} ,
    type        = abbrev,
    parent      = prob,
}

\newglossaryentry{causal}{type=abbrev,name=Causality,description={\nopostdesc}}

\newglossaryentry{cause}{
    name        = \ensuremath{\boldsymbol{N}},
    description = {noise (independent)  variable vector} ,
    type        = abbrev,
    parent      = causal,
}

\newglossaryentry{causecomp}{
    name        = \ensuremath{N},
    description = {noise (independent)  variable component} ,
    type        = abbrev,
    parent      = causal,
}

\newglossaryentry{Cause}{
    name        = \ensuremath{\mathcal{N}} ,
    description = {space of the noise variables} ,
    type        = abbrev,
    parent      = causal,
}

\newglossaryentry{effect}{
    name        = \ensuremath{\boldsymbol{X}},
    description = {observation vector} ,
    type        = abbrev,
    parent      = causal,
}
\newglossaryentry{effectcomp}{
    name        = \ensuremath{X},
    description = {observation component} ,
    type        = abbrev,
    parent      = causal,
}

\newglossaryentry{Effect}{
    name        = \ensuremath{\mathcal{X}} ,
    description = {space of the effect variables} ,
    type        = abbrev,
    parent      = causal,
}

\newglossaryentry{pa}{
    name        = \ensuremath{\boldsymbol{Pa}},
    description = {parents of \gls{effect}} ,
    type        = abbrev,
    parent      = causal,
}

\newglossaryentry{nondesc}{
    name        = \ensuremath{\boldsymbol{ND}},
    description = {non-descendants of \gls{effect}} ,
    type        = abbrev,
    parent      = causal,
}

\newglossaryentry{nondescminuspa}{
    name        = \ensuremath{\boldsymbol{\overline{ND}}},
    description = {non-descendants of \gls{effect}, excluding its parents} ,
    type        = abbrev,
    parent      = causal,
}

\newglossaryentry{semf}{
    name        = \ensuremath{\boldsymbol{f}},
    description = {structural assignment in \glspl{sem}} ,
    type        = abbrev,
    parent      = causal,
}

\newglossaryentry{semfcomp}{
    name        = \ensuremath{f},
    description = {a component of \gls{semf}} ,
    type        = abbrev,
    parent      = causal,
}

\newglossaryentry{order}{
    name        = \ensuremath{\pi},
    description = {causal ordering} ,
    type        = abbrev,
    parent      = causal,
}

\newglossaryentry{indexset}{
    name        = \ensuremath{\mathcal{I}},
    description = {index set} ,
    type        = abbrev,
    parent      = causal,
}
\newglossaryentry{adjacency}{
    name        = \ensuremath{\boldsymbol{\mathcal{A}}} ,
    description = {adjacency matrix of a \glspl{sem}} ,
    type        = abbrev,
    parent      = causal,
}

\newglossaryentry{connectivity}{
    name        = \ensuremath{\boldsymbol{\mathcal{C}}} ,
    description = {connectivity matrix of a \glspl{sem}} ,
    type        = abbrev,
    parent      = causal,
}

\newglossaryentry{dependency}{
    name        = \ensuremath{\mathcal{D}} ,
    description = {dependency matrix of a \glspl{sem}} ,
    type        = abbrev,
    parent      = causal,
}

\newglossaryentry{seq}{
    name        = \ensuremath{\sim_{\acrshort{dag}}} ,
    description = {structural equivalence} ,
    type        = abbrev,
    parent      = causal,
}

\newglossaryentry{contrastive}{type=abbrev,name=Contrastive Learning,description={\nopostdesc}}
\newglossaryentry{clloss}{
    name        = \ensuremath{\mathcal{L}_{\mathrm{\acrshort{cl}}}} ,
    description = {contrastive loss function} ,
    type        = abbrev,
    parent      = contrastive,
}

\newglossaryentry{alignloss}{
    name        = \ensuremath{\mathcal{L}_{\mathrm{align}}} ,
    description = {alignment term in \gls{clloss}} ,
    type        = abbrev,
    parent      = contrastive,
}

\newglossaryentry{uniformloss}{
    name        = \ensuremath{\mathcal{L}_{\mathrm{uniform}}} ,
    description = {uniformity term in \gls{clloss}} ,
    type        = abbrev,
    parent      = contrastive,
}

\newglossaryentry{temp}{
    name        = \ensuremath{{\boldsymbol{\tau}}} ,
    description = {temperature in \gls{clloss}} ,
    type        = abbrev,
    parent      = contrastive,
}

\newglossaryentry{numneg}{
    name        = \ensuremath{M} ,
    description = {number of negative samples} ,
    type        = abbrev,
    parent      = contrastive,
}

\newglossaryentry{vaes}{type=abbrev,name=\acrlongpl{vae},description={\nopostdesc}}

\newglossaryentry{q}{
    name        = \ensuremath{q_{\gls{encpar}}(\gls{latent}|\gls{obs})} ,
    description = {variational posterior of the \acrshort{vae}, mapping $\gls{obs}\mapsto\gls{latent}$ parametrized by \gls{encpar}} ,
    type        = abbrev,
    parent      = vaes,
}
\newglossaryentry{qopt}{
    name        = \ensuremath{q_{\widehat{\gls{encpar}}}(\gls{latent}|\gls{obs})} ,
    description = {optimal variational posterior of the \acrshort{vae}, mapping $\gls{obs}\mapsto\gls{latent}$ parametrized by \gls{encpar}} ,
    type        = abbrev,
    parent      = vaes,
}

\newglossaryentry{encpar}{
    name        = \ensuremath{\boldsymbol{\phi}} ,
    description = {parameters of the variational posterior \gls{q}} ,
    type        = abbrev,
    parent      = vaes,
}

\newglossaryentry{encparopt}{
    name        = \ensuremath{\widehat{\boldsymbol{\phi}}} ,
    description = {optimal parameters of the variational posterior \gls{q}} ,
    type        = abbrev,
    parent      = vaes,
}

\newglossaryentry{var_family}{
    name        = \ensuremath{\mathcal{Q}} ,
    description = {distribution family of the variational posterior \gls{q} } ,
    type        = abbrev,
    parent      = vaes,
}

\newglossaryentry{pz}{
    name        = \ensuremath{p_0(\gls{latent})} ,
    description = {latent prior distribution} ,
    type        = abbrev,
    parent      = vaes,
}

\newglossaryentry{px}{
    name        = \ensuremath{p_{\gls{decpar}}(\gls{obs})} ,
    description = {marginal likelihood } ,
    type        = abbrev,
    parent      = vaes,
}

\newglossaryentry{pdata}{
    name        = \ensuremath{p(\gls{obs})} ,
    description = {data distribution } ,
    type        = abbrev,
    parent      = vaes,
}

\newglossaryentry{mean_enc}{
    name        = \ensuremath{\mu_{\gls{latent}|\gls{obs}}} ,
    description = {mean encoder of the \acrshort{vae}, \ie, $\expectation{\gls{latent}\sim\gls{q}}\parenthesis{\gls{latent}}$, mapping $\gls{obs}\mapsto\gls{latent}$} ,
    type        = abbrev,
    parent      = vaes,
}

\newglossaryentry{var_cov}{
    name        = \ensuremath{\gls{cov}^{\gls{encpar}}_{\gls{latent}|\gls{obs}}} ,
    description = {covariance matrix of \gls{q}} ,
    type        = abbrev,
    parent      = vaes,
}

\newglossaryentry{sigmak}{
    name        = \ensuremath{{\sigma}_{k}^{\gls{encpar}}(\gls{obs})^{2}} ,
    description = {variance of \gls{q} in dimension $k$} ,
    type        = abbrev,
    parent      = vaes,
}

\newglossaryentry{sigmaopt}{
    name        = \ensuremath{\boldsymbol{\sigma}^{\gls{encparopt}}(\gls{obs})^{2}} ,
    description = {optimal variance of \gls{q}} ,
    type        = abbrev,
    parent      = vaes,
}

\newglossaryentry{sigmaoptk}{
    name        = \ensuremath{{\sigma}_{k}^{\gls{encparopt}}(\gls{obs})^{2}} ,
    description = {optimal variance of \gls{q} in dimension $k$} ,
    type        = abbrev,
    parent      = vaes,
}

\newglossaryentry{mu}{
    name        = \ensuremath{\boldsymbol{\mu}^{\gls{encpar}}(\gls{obs})} ,
    description = {mean of \gls{q}} ,
    type        = abbrev,
    parent      = vaes,
}

\newglossaryentry{muk}{
    name        = \ensuremath{{\mu}_{k}^{\gls{encpar}}(\gls{obs})} ,
    description = {mean of \gls{q} in dimension $k$} ,
    type        = abbrev,
    parent      = vaes,
}

\newglossaryentry{muopt}{
    name        = \ensuremath{\boldsymbol{\mu}^{\gls{encparopt}}(\gls{obs})} ,
    description = {optimal mean of \gls{q}} ,
    type        = abbrev,
    parent      = vaes,
}

\newglossaryentry{muoptk}{
    name        = \ensuremath{{\mu}_{k}^{\gls{encparopt}}(\gls{obs})} ,
    description = {optimal mean of \gls{q} in dimension $k$} ,
    type        = abbrev,
    parent      = vaes,
}

\newglossaryentry{gamma}{
    name        = \ensuremath{\gamma} ,
    description = {square root of the precision of the \gls{vae} decoder} ,
    type        = abbrev,
    parent      = vaes,
}

\newglossaryentry{betaloss}{
    name        = \ensuremath{\mathcal{L}_{\beta}} ,
    description = {\betavae loss function} ,
    type        = abbrev,
    parent      = vaes,
}

\newglossaryentry{pxz}{
    name        = \ensuremath{p_{\gls{decpar}}(\gls{obs}|\gls{latent})} ,
    description = {conditional distribution of the decoded samples of the \acrshort{vae}, mapping $\gls{latent}\mapsto\gls{obs}$, parametrized by \gls{decpar}} ,
    type        = abbrev,
    parent      = vaes,
}
\newglossaryentry{pzx}{
    name        = \ensuremath{p_{\gls{decpar}}(\gls{latent}|\gls{obs})} ,
    description = {true posterior distribution of the decoded samples of the \acrshort{vae}, mapping $\gls{obs}\mapsto\gls{latent}$, parametrized by \gls{decpar}} ,
    type        = abbrev,
    parent      = vaes,
}
\newglossaryentry{decpar}{
    name        = \ensuremath{\boldsymbol{\theta}} ,
    description = {parameters of the decoder \gls{pxz}} ,
    type        = abbrev,
    parent      = vaes,
}

\newglossaryentry{invdeccomp}{
    name        = \ensuremath{{g}^{\gls{decpar}}} ,
    description = {inverse decoder component} ,
    type        = abbrev,
    parent      = vaes,
}

\newglossaryentry{invdec}{
    name        = \ensuremath{\mathrm{\boldsymbol{g}}^{\gls{decpar}}} ,
    description = {inverse decoder} ,
    type        = abbrev,
    parent      = vaes,
}

\newglossaryentry{distortion}{
    name        = \ensuremath{D} ,
    description = {Distortion of \cite{alemi_fixing_2018}, the same as the reconstruction term of the \acrshort{elbo} for $\beta=1$} ,
    type        = abbrev,
    parent      = vaes,
}

\newglossaryentry{rate}{
    name        = \ensuremath{R} ,
    description = {Rate of \cite{alemi_fixing_2018}, the same as the \acrshort{kld} term of the \acrshort{elbo} for $\beta=1$} ,
    type        = abbrev,
    parent      = vaes,
}

\newglossaryentry{lindec}{
    name        = \ensuremath{\boldsymbol{\mathrm{W}}} ,
    description = {weight matrix of a linear decoder} ,
    type        = abbrev,
    parent      = vaes,
}

\newglossaryentry{linenc}{
    name        = \ensuremath{\boldsymbol{\mathrm{V}}} ,
    description = {weight matrix of a linear encoder} ,
    type        = abbrev,
    parent      = vaes,
}

\newglossaryentry{imas}{type=abbrev,name=\acrlong{ima},description={\nopostdesc}}

\newglossaryentry{mixing}{
    name        = \ensuremath{\inv{g}} ,
    description = {inverse of the learned unmixing of the \acrshort{ima}, mapping $\gls{latent}\mapsto\gls{obs}$ } ,
    type        = abbrev,
    parent      = imas,
}

\newglossaryentry{lin_mixing}{
    name        = \ensuremath{A} ,
    description = {ground-truth \emph{linear} mixing process of the \acrshort{ima}, mapping $\gls{latent}\mapsto\gls{obs}$ } ,
    type        = abbrev,
    parent      = imas,
}

\newglossaryentry{cima_local}{
    name        = \ensuremath{c_{\acrshort{ima}}} ,
    description = {local \acrshort{ima} contrast } ,
    type        = abbrev,
    parent      = imas,
}

\newglossaryentry{cima_global}{
    name        = \ensuremath{C_{\acrshort{ima}}} ,
    description = {global \acrshort{ima} contrast } ,
    type        = abbrev,
    parent      = imas,
}

\newglossaryentry{source}{
    name        = \ensuremath{s} ,
    description = {sources (\acrshort{ica} equivalent of latents)} ,
    type        = abbrev,
    parent      = imas,
}

\newglossaryentry{rec_s}{
    name        = \ensuremath{\boldsymbol{y}} ,
    description = {reconstructed sources} ,
    type        = abbrev,
    parent      = imas,
}

\newglossaryentry{p_source}{
    name        = \ensuremath{p_{\gls{latent}}} ,
    description = {source distribution} ,
    type        = abbrev,
    parent      = imas,
}

\newglossaryentry{imaloss}{
    name        = \ensuremath{\mathcal{L}_{\gls{ima}}} ,
    description = {\gls{ima} loss function} ,
    type        = abbrev,
    parent      = imas,
}

\NewDocumentCommand{\cima}{ O{\gls{dec}} O{\gls{latent}}  }{\ensuremath{\gls{cima_local} ( #1\!,  #2) }\xspace}
\NewDocumentCommand{\Cima}{ O{\gls{dec}} O{\ensuremath{p_0}  }}{\ensuremath{\gls{cima_global} ( #1,  #2) }\xspace}

\newglossaryentry{gps}{type=abbrev,name=\acrlongpl{gp},description={\nopostdesc}}
\newglossaryentry{gpr}{
    name        = \ensuremath{\mathcal{GP}} ,
    description = {Gaussian Process} ,
    type        = abbrev,
    parent      = gps,
}

\newglossaryentry{gpker}{
    name        = \ensuremath{k} ,
    description = {kernel function} ,
    type        = abbrev,
    parent      = gps,
}

\newglossaryentry{gpcov}{
    name        = \ensuremath{\mathcal{K}} ,
    description = {$\gls{numdata}\times\gls{numdata}$ covariance matrix of a \acrshort{gp}} ,
    type        = abbrev,
    parent      = gps,
}

\makeglossaries

%% file: acronyms.tex
\newacronym{mpa}{MPA}{Measure Preserving Automorphism}
\newacronym{iid}{i.i.d.}{independent and identically distributed}
\newacronym{vmf}{vMF}{von Mises-Fisher}
\newacronym{nivmf}{nivMF}{non-isotropic von Mises-Fisher}
\newacronym{pd}{PD}{positive definite}
\newacronym{psd}{PSD}{positive semi-definite}
\newacronym{nd}{ND}{negative definite}
\newacronym{nsd}{NSD}{negative semi-definite}

\newacronym{ode}{ODE}{Ordinary Differential Equation}
\newacronym{pde}{PDE}{Partial Differential Equation}

\newacronym{lhs}{LHS}{left hand side}
\newacronym{rhs}{RHS}{right hand side}

\newacronym{rv}{RV}{random variable}

\newacronym{ae}{AE}{AutoEncoder}
\newacronym{lae}{LAE}{Linear Autoencoder}
\newacronym{vae}{VAE}{Variational Autoencoder}
\newacronym{cvvae}{CV-VAE}{Constant-Variance Variational Autoencoder}
\newacronym{ivae}{iVAE}{Identifiable Variational Autoencoder}
\newacronym{rae}{RAE}{Regularized Autoencoder}
\newacronym{grae}{GRAE}{Gaussian Regularized Autoencoder}
\newacronym{lvm}{LVM}{latent variable model}

\newacronym[longplural=Gaussian Processes]{gp}{GP}{Gaussian Process}
\newacronym{gplvm}{GPLVM}{Gaussian Process Latent Variable Model}
\newacronym{rbf}{RBF}{Radial Basis Function}

\newcommand{\betavae}{$\beta$-\gls{vae}\xspace}

\newacronym{kld}{KL}{Kullback-Leibler Divergence}
\newacronym{elbo}{{\text{\upshape ELBO}}}{evidence lower bound}
\newacronym{pca}{PCA}{Principal Component Analysis}
\newacronym{ppca}{PPCA}{Probabilistic Principal Component Analysis}
\newacronym{ebm}{EBM}{Energy-Based Model}
\newacronym{cca}{CCA}{Canonical Correlation Analysis}

\newacronym{mi}{MI}{Mutual Information}

\newacronym[longplural=Identifiable Exchangeable Mechanisms]{iem}{IEM}{Identifiable Exchangeable Mechanisms}
\newacronym[longplural=Independent Causal Mechanisms]{icm}{ICM}{Independent Causal Mechanisms}
\newacronym{sms}{SMS}{Sparse Mechanism Shift}
\newacronym{mss}{MSS}{Mechanism Shift Score}
\newacronym{sem}{SEM}{Structural Equation Model}
\newacronym{lingam}{LiNGAM}{Linear Non-Gaussian Acyclic Model}
\newacronym{dag}{DAG}{Directed Acyclic Graph}
\newacronym{anm}{ANM}{Additive Noise Model}
\newacronym{cd}{CD}{Causal Discovery}
\newacronym{crl}{CRL}{Causal Representation Learning}
\newacronym{hmm}{HMM}{Hidden Markov Model}
\newacronym{plr}{PLR}{Partially Linear Regression}

\newacronym{it}{IT}{identifiability theory}
\newacronym{sith}{SITh}{Singular Identifiability Theory}

\newacronym{lt}{LT}{learning theory}
\newacronym{slt}{SLT}{Singular Learning Theory}

\newacronym{ica}{ICA}{Independent Component Analysis}
\newacronym{nlica}{NLICA}{nonlinear Independent Component Analysis}
\newacronym{bss}{BSS}{Blind Source Separation}

\newacronym{ima}{{\text{\upshape IMA}}}{Independent Mechanism Analysis}
\newacronym{igci}{IGCI}{Information Geometric Causal Inference}
\newacronym{cdf}{CdF}{Causal de Finetti}

\newacronym{nce}{NCE}{Noise Contrastive Estimation}
\newacronym{pcl}{PCL}{Permutation-Contrastive Learning}
\newacronym{tcl}{TCL}{Time-Contrastive Learning}
\newacronym{gencl}{GCL}{Generalized Contrastive Learning}
\newacronym{iia}{IIA}{Independent Innovation Analysis}

\newacronym{ar}{AR}{autoregressive}
\newacronym{var}{VAR}{Vector autoregressive}
\newacronym{nvar}{NVAR}{Nonlinear Vector AutoRegressive}

\newacronym{ai}{AI}{Artificial Intelligence}
\newacronym{ml}{ML}{Machine Learning}
\newacronym{dml}{DML}{Double Machine Learning}
\newacronym{homl}{HOML}{Higher-order Orthogonal Machine Learning}
\newacronym{dl}{DL}{Deep Learning}
\newacronym{rl}{RL}{Reinforcement Learning}
\newacronym{mbrl}{MBRL}{Model-Based Reinforcement Learning}
\newacronym{rlhf}{RLHF}{Reinforcement Learning from Human Feedback}
\newacronym{ssl}{SSL}{self-supervised learning}

\newacronym{cl}{CL}{Contrastive Learning}
\newacronym{dcl}{DCL}{Debiased Contrastive Learning}
\newacronym{scl}{SCL}{Spectral Contrastive Learning}
\newacronym{gcl}{GCL}{Graph Contrastive Learning}
\newacronym{alphacl}{$\alpha$-CL}{$\alpha$-Contrastive Learning}
\newacronym{arcl}{ArCL}{Augmentation-robust Contrastive Learning}
\newacronym{fce}{FCE}{Flow Contrastive Estimation}
\newacronym{pid}{PID}{parametric instance discrimination}

\newacronym{vince}{VINCE}{Variational InfoNCE}
\newacronym{rince}{RINCE}{Robust InfoNCE}
\newacronym{aggnce}{AggNCE}{Aggregated InfoNCE}
\newacronym{mcinfonce}{MCInfoNCE}{Monte-Carlo InfoNCE}

\newacronym{gmc}{GMC}{Geometric Multimodal Contrastive Learning}
\newacronym{looc}{LooC}{Leave-one-out Contrastive Learning}
\newacronym{npc}{NPC}{Negative-Positive Coupling}
\newacronym{cpc}{CPC}{Contrastive Predictive Coding}
\newacronym{nlp}{NLP}{Natural Language Processing}
\newacronym{gdl}{GDL}{Geometric Deep Learning}
\newacronym{msn}{MSN}{Masked Siamese Networks}
\newacronym{ifm}{IFM}{Implicit Feature Modification}

\newacronym{dnn}{DNN}{Deep Neural Network}
\newacronym{nn}{NN}{Neural Network}
\newacronym{ann}{ANN}{Artificial Neural Network}

\newacronym{fm}{FM}{Foundation Model}
\newacronym{llm}{LLM}{Large Language Model}
\newacronym{vlm}{VLM}{Vision-Language Model}
\newacronym{pcfg}{PCFG}{Probabilistic Context-Free Grammar}
\newacronym{icl}{ICL}{in-context learning}

\newacronym{nc}{NC}{Neural Collapse}
\newacronym{cdt}{CDT}{Class-Dependent Temperature}
\newacronym{mlp}{MLP}{Multi-Layer Perceptron}
\newacronym{fc}{FC}{Fully Connected}
\newacronym{strnn}{StrNN}{Structured Neural Network}

\newacronym{cn}{conv}{Convolutional layer}
\newacronym{cnn}{CNN}{Convolutional Neural Network}
\newacronym{gnn}{GNN}{Graph Neural Network}
\newacronym{ssm}{SSM}{State Space Model}

\newacronym{rnn}{RNN}{Recurrent Neural Network}
\newacronym{lstm}{LSTM}{Long Short-Term Memory}
\newacronym{gru}{GRU}{Gated Recurrent Unit}
\newacronym{relu}{ReLU}{Rectified Linear Unit}
\newacronym{bn}{BN}{Batch Normalization}
\newacronym{dbn}{DBN}{Decorrelated Batch Normalization}

\newacronym{gan}{GAN}{Generative Adversarial Network}

\newacronym{diayn}{DIAYN}{Diversity Is All You Need}
\newacronym{dads}{DADS}{DYnamics-Aware Discovery of Skills}
\newacronym{sac}{SAC}{Soft Actor Critic}
\newacronym{a2c}{A2C}{Advantage Actor Critic}

\newacronym{sgd}{SGD}{Stochastic Gradient Descent}
\newacronym{adam}{ADAM}{Adaptive Moment Estimation}
\newacronym{svd}{SVD}{Singular Value Decomposition}
\newacronym{wls}{WLS}{Weighted Least Squares}

\newacronym{sam}{SAM}{Sharpness-Aware Minimization}
\newacronym{samba}{SAMBA}{SAM-Based Autoencoder}
\newacronym{vi}{VI}{Variational Inference}
\newacronym{mfvi}{MFVI}{Mean Field Variational Inference}

\newacronym[longplural=data generating processes]{dgp}{DGP}{data generating process}
\newacronym{map}{MAP}{Maximum A Posteriori}
\newacronym{mle}{MLE}{maximum likelihood estimation}

\newacronym{etf}{ETF}{Equiangular Tight Frame}

\newacronym{mse}{MSE}{Mean Squared Error}
\newacronym{mae}{MAE}{Mean Absolute Error}
\newacronym{ce}{{\text{\upshape CE}}}{cross entropy}

\newacronym{sid}{SID}{Structural Intervention Distance}
\newacronym{shd}{SHD}{Structural Hamming Distance}

\newacronym{mcc}{MCC}{Mean Correlation Coefficient}
\newacronym{mig}{MIG}{Mutual Information Gap}
\newacronym{dci}{DCI}{Disentanglement Completeness Informativeness score}

\newacronym{arc}{ARC}{Average Relative Confusion}
\newacronym{acr}{ACR}{Average Confusion Ratio}

\newacronym{api}{API}{Application Programming Interface}
\newacronym{cpu}{CPU}{Central Processing Unit}
\newacronym{gpu}{GPU}{Graphics Processing Unit}

\newacronym{lti}{LTI}{Linear Time-Invariant}
\newacronym{zoh}{ZOH}{Zero-Order Hold}

\newacronym{gt}{{\text{\upshape GT}}}{ground truth}
\newacronym{ood}{OOD}{out-of-distribution}
\newacronym{oov}{OOV}{out-of-variablme}
\newacronym{fsm}{FSM}{Finite State Machine}
\newacronym{rasp}{RASP}{Restricted-Access Sequence Processing Language}

\newacronym{ntk}{NTK}{Neural Tangent Kernel}

\newacronym{as}{a.s.}{almost surely}
\newacronym{alev}{a.e.}{almost everywhere}

\newacronym{sos}{SOS}{start-of-sequence}
\newacronym{eos}{EOS}{end-of-sequence}

\newacronym{prh}{PRH}{Platonic Representation Hypothesis}

%% file: abstract.tex
Self-Supervised Learning (SSL) powers many current AI systems. As research interest and investment grow, the SSL design space continues to expand. The Platonic view of SSL, following the \emph{Platonic Representation Hypothesis} (PRH), suggests that despite different methods and engineering approaches, all representations converge to the same Platonic ideal. However, this phenomenon lacks precise theoretical explanation. By synthesizing evidence from Identifiability Theory (IT), we show that the PRH can emerge in SSL.
There is a gap between SSL theory and practice: Current IT cannot explain SSL's empirical success, though it has practically relevant insights. Our work formulates a blueprint for SSL research to bridge this gap: we propose expanding IT into what we term Singular Identifiability Theory (SITh), a broader theoretical framework encompassing the entire SSL pipeline. SITh would allow deeper insights into the implicit data assumptions in SSL and advance the field towards learning more interpretable and generalizable representations. We highlight three critical directions for future research: 1) training dynamics and convergence properties of SSL; 2) the impact of finite samples, batch size, and data diversity; and 3) the role of inductive biases in architecture, augmentations, initialization schemes, and optimizers.

%% file: main_text.tex
%

\section{Introduction}

\Gls{ssl} drives many current AI breakthroughs in language, vision, and image-text models~\citep{radford_language_2018,radford2021learningtransferablevisualmodels,assran_self-supervised_2023}. This remarkable success builds on a plethora of algorithms and engineering practices. However, the model zoo makes \gls{ssl} hard to navigate, divides the attention for a better theoretical understanding and, we argue, dampens future advancements. 
\gls{ssl} is likely to continue being a foundational paradigm of machine learning, but it needs to be consolidated, similar to the program of unifying deep learning through geometry \cite{bronstein2021geometric}. There were recent attempts to unify the \gls{ssl} model zoo~\citep{wang_understanding_2020,morningstar_augmentations_2024,bizeul_probabilistic_2024,fleissner_probabilistic_2025}. \citet{huh_platonic_2024} even argued, by positing the \gls{prh}, that neural networks converge to effectively the same representation, despite different setups, loss functions, data modalities, and engineering practices. However, their analysis does not provide a conclusive mathematical answer \textit{when and why} this happens.
We do have theories that can provide such an answer, but a gap exists between \gls{ssl} theory and practice. Prevalent frameworks such as \gls{it} cannot explain the role of initialization, stop gradients, or learning dynamics, and neither can solve problems like dimensional collapse \cite{jing_understanding_2022} or loss saturation.
An even larger problem of current theory is generalization:
The promise of self-supervised representations to work well on many downstream tasks alludes to ``universality," or at least \gls{ood} generalization. However, 
current theories mostly focus on the \acrshort{iid} case, and empirical methods generally perform well only in-distribution~\citep{montero_role_2021,schott_visual_2021,montero_lost_2024, mayilvahanan_does_2024,mayilvahanan_search_2024}.

        \begin{tcolorbox}[colback=figgreen!15!white,colframe=figgreen!75!black, title={\textbf{Position:} empirical advancements alone are insufficient to accelerate SSL research, we need to bridge the gap between theory and practice.}]
           We need an empirically grounded theory, which we call \acrfull{sith}, to accelerate SSL research by: 
                \begin{enumerate}[nolistsep, leftmargin=*, label=(\roman*)]
                    \item Designing realistic \glspl{dgp}, grounded in empirical observations;
                    \item Formalizing when and why self-supervised representations are converging;
                    \item Providing principled recommendations for designing and evaluating \gls{ssl}.
                \end{enumerate}
        \end{tcolorbox}

\vspace{-.5em}

\textbf{This position paper argues that to accelerate SSL research, we need to develop a theoretical framework grounded in empirical observation.} Acknowledging the success of \gls{it}, we call for its extension to what we term \gls{sith}---alluding to \gls{slt}~\citep{watanabe_algebraic_2009,watanabe_mathematical_2020}, which extended \gls{lt} to considerations relevant to neural networks. 
SSL practitioners need to rely on principled guidance derived from theory, whereas theoreticians need to concentrate on studying practically relevant aspects of \gls{ssl}---we provide concrete examples and research questions in \cref{table:blind_spots}.
Our \textbf{contributions} are:
\vspace{-.3em}
\begin{itemize}[nolistsep, leftmargin=*]
    \item We provide an approachable introduction to \acrfull{it} for practitioners, demonstrating how SSL practice can benefit from \gls{it} (\cref{sec:it});
    \item We call for an empirically grounded extension of identifiability, termed  \acrfull{sith} to close the gap between SSL theory and practice, mainly relying on the concept of a \gls{dgp} (\cref{sec:pos});
    \item We provide an extensive overview of the open questions in SSL, formulate concrete research questions and promising research directions to suggest how empirical observations and theoretical results can drive \gls{ssl} forward (\cref{table:blind_spots}).
\end{itemize}
\vspace{-.4em}

\section{Background}

    \paragraph{\gls{ssl} drives most AI breakthroughs.}
        Long gone are the days when supervised, semi-supervised, and unsupervised learning co-existed as the main tracks of top AI conferences. The emergence of Internet-scale datasets coupled with the need for foundation models \cite{bommasani2021opportunities}---models capable of solving many tasks post-training---has moved the landscape of AI under one umbrella: \acrfull{ssl}. With access to data and compute, hundreds of SSL methods have emerged~\citep{balestriero_cookbook_2023}. As a result, SSL now holds the state-of-the-art across numerous domains like remote sensing and medical data \cite{wang2022self,krishnan2022self}, and applications such as zero-shot reasoning \cite{radford2021learningtransferablevisualmodels,zhai2023sigmoid}, multi-modal learning \cite{girdhar2022omnivore}.
        The distinct engineering choices and implementation details make it hard to empirically test every hypothesis one may have in mind. Yet, such hypothesis testing is becoming more important than ever to understand possible robustness, fairness, compositional, or structural limitations of existing methods. Without a clear understanding of today's limitations, it is getting harder to develop tomorrow's methods.

    \vspace{-.5em}
    \paragraph{Self-supervised representations are often but not always similar.}

        Theoretical and empirical evidence indicates a surprising similarity between some self-supervised representations. \citet{roeder_linear_2020} proved that in specific cases, two trained models learn a similar representation up to a linear representation.
        Works on relative representations and model stitching~\citep{moschella_relative_2023,cannistraci_bricks_2023,norelli_asif_2023,fumero_latent_2024,maiorca_latent_2024} demonstrated that learned self-supervised representations can be related via simple (\eg, linear) transformations. A similar notion of representation linearity was demonstrated in language models~\citep{park_linear_2023,marconato_all_2024,jha_harnessing_2025}, building from earlier observations of analogy making in word embeddings like \texttt{word2vec}~\citep{mikolov_distributed_2013,arora_latent_2016}.
        The \gls{prh}~\citep{huh_platonic_2024} proposed an explanation of why such convergence can happen, alluding to Plato's allegory of the cave. \citet{huh_platonic_2024} formalized three hypotheses, corresponding to the effect of: i) \textit{function class}, ii) \textit{data sets and loss functions}, and iii) \textit{regularization}. 
        Earlier work by \citet{duan_unsupervised_2020} suggested a link between model convergence and identifiability.
        We discuss evidence for their importance in \gls{ssl} and highlight that current theories cannot fully explain their effect (\cf \cref{sec:prh} for details). \citet{morningstar_augmentations_2024} concluded that the data augmentations matter more than the \gls{ssl} method for downstream classification accuracy on ImageNet~\citep{krizhevsky2012imagenet}. \citet{ciernik_training_2024} demonstrated that \gls{ssl} representations are often, but not universally, similar (\cf \cref{fig:similarity_matrix_cka_kernel_rbf}). The prevalent problems of dimensional collapse \cite{jing_understanding_2022} or the projector phenomenon~\citep{jing_understanding_2022,chen_simple_2020,von_kugelgen_self-supervised_2021}, suggest that the \gls{prh} can only be true for a set of models.
        Thus, a mathematically conclusive answer as of \textit{when and why} the \gls{prh} can hold (in \gls{ssl}) eludes the research community. Our work aims to guide research interest towards answering these questions.

\section{\Acrlong{it}}\label{sec:it}
    This section provides an intuitive introduction to the terminology of \acrfull{it} (\cref{subsec:it_terminology}), and demonstrates IT's merits for practitioners on the example of SimCLR (\cref{subsec:it_simclr}). We also review the state of the field (\cref{subsec:it_bg}); and highlight many \gls{it}-driven applications~(\cref{subsec:it_applications}).

     \subsection{The terminology of identifiability}\label{subsec:it_terminology}

Consider a diverse collection of images like ImageNet. For each image, certain ``ingredients'' or latents came together to create it: the type of object, its position, color, lighting conditions, and so on. \Acrfull{it} provides a mathematical framework to study whether we can discover these underlying properties just by looking at the images.

Imagine these images were created by a virtual ``renderer''---what IT calls a \textit{\acrfull{dgp}}---that takes these latents as input and produces the final image~\cite{kulkarni2015deep}. For instance, to generate a photo of a red car, this renderer would need inputs specifying ``car'' as the object class, ``red'' for its color, and values for its position. The key question IT asks is: 
\begin{center}
\textit{Given only the images, under what conditions can we work backwards to uncover the original latents through \gls{ssl}?}
\end{center}
This is not always straightforward since there might be multiple valid ways to describe the same latent. Take color: we could represent it using RGB values or HSV coordinates—both produce identical images~\cite{higgins_towards_2018}. IT handles this by grouping these equivalent descriptions into \textit{equivalence classes}. To pin down a specific solution, we need additional assumptions about the DGP, e.g., regarding the distribution of latents or the function class of the renderer~\cite{sprekeler2014extension}. %

    \subsection{What can identifiability bring to SSL? A gentle introduction with SimCLR}\label{subsec:it_simclr}
        To illustrate how \gls{it} can improve our understanding of SSL methods, we consider SimCLR~\citep{chen_simple_2020}. 
        \paragraph{Step 1: the birth of SimCLR.} SimCLR was proposed from an empirical perspective without any identifiability theoretic considerations. Its success inspired theoreticians to try to understand its operating principles. 
        \paragraph{Step 2: Uniformity and alignment.} The first theoretical result by \citet{wang_understanding_2020} conceptualized the SimCLR loss in terms of two opposing ``forces:" an attractive force termed \textit{alignment} pulling the representations of similar (augmented, positive) samples together, whereas a repelling force termed \textit{uniformity} pushing dissimilar (negative) samples apart. This conceptual framework provided some practical guidance on design choices, including the temperature value or batch size. However, problems such as dimensional collapse~\cite{jing_understanding_2022} (when some latent information is not learned), or the emergence of the projector---when downstream performance is better for the non-ultimate layer---remained unexplained.
        \paragraph{Step 3: Identifiable DGP.} \citet{zimmermann_contrastive_2021} showed---building on prior works in nonlinear \gls{ica}~\citep{hyvarinen_unsupervised_2016,hyvarinen_nonlinear_2019,khemakhem_variational_2020,khemakhem_ice-beem_2020,hyvarinen_nonlinear_2017}---that the SimCLR objective is optimized when the neural network learns to parametrize a DGP on the hypersphere, where the augmentations are modeled with a \gls{vmf} conditional distribution. This result showed how a few assumptions are sufficient for identifiability: a hypersphere latent space with uniformly distributed samples, augmentation distributions according to a \gls{vmf}, and an invertible encoder. 
        Furthermore, this result made the \textit{implict} assumptions about the data and the underlying DGP explicit and pinpointed their unrealistic nature: the \gls{vmf} conditional is isotropic, which does not correspond to realistic augmentations, including large crops, which are crucial for SSL.
        Thus, IT can model \textit{simplified} SSL methods, but not the practically used versions.
        \paragraph{Step 4: towards a more realistic theory.} A later extension by \citet{rusak_infonce_2024} proved identifiability for a conditional which can model augmentations with different strengths. Yet, we still do not understand (and cannot mitigate) dimensional collapse~\cite{jing_understanding_2022} or the projector.

    \subsection{The state of identifiability theory}\label{subsec:it_bg}

        The prevalent method family with identifiability guarantees is \acrfull{ica}~\citep{comon1994independent,hyvarinen_nonlinear_1999,hyvarinen_independent_2001,hyvarinen_identifiability_2023}. \gls{ica} methods aim to extract (conditionally) independent latent variables, called sources---this problem is also called \gls{bss}. 
        Though identifiability is impossible in the nonlinear case without further assumptions~\citep{hyvarinen_nonlinear_1999,locatello_challenging_2019}, recent nonlinear \gls{ica} methods demonstrated identifiability results for increasingly realistic settings, pushing forward our understanding of when, how, and why \gls{ssl} could work~\citep{khemakhem_ice-beem_2020,khemakhem_variational_2020,reizinger_embrace_2022,gresele_independent_2021,morioka_independent_2021,morioka_connectivity-contrastive_2023,hyvarinen_unsupervised_2016,hyvarinen_nonlinear_2019,klindt_towards_2021}.
        Identifiability guarantees are also central in the field of causal inference~\citep{pearl_causality_2009}. Causal methods are promising data efficiency and generalization, thus, they are of interest for practice. The new field of \gls{crl}~\citep{scholkopf_towards_2021} also has identifiability guarantees~\citep{von_kugelgen_nonparametric_2023,von_kugelgen_identifiable_2024,wendong_causal_2023,brehmer_weakly_2022,ahuja_interventional_2022}, some of which were shown to be deeply connected to \gls{ica} methods~\citep{reizinger_identifiable_2024,hyvarinen_identifiability_2023}. 
        Recently, \citet{reizinger_cross-entropy_2024} showed that even for standard classification tasks, a simple \gls{dgp} can be formulated that enjoys linear identifiability. Moreover, this is easily estimated by maximum likelihood cross-entropy minimization, generalizing the result of \citet{roeder_linear_2020} and providing evidence about why \citet{huh_platonic_2024} observed representational similarity across many training objectives---namely, they were different formulations of cross-entropy minimization. 
        Despite delivering insights about widely-used \gls{ssl} methods~\citep{zimmermann_contrastive_2021}, most theoretical works are fairly idealistic as they make strong assumptions, including assuming infinite data and infinite training time (\ie, model convergence).

    \subsection{Applications: why practitioners should care}\label{subsec:it_applications}
        The biggest critique of \gls{it}---and machine learning theories in general---is that 1) most progress in the field comes from empirical techniques or scaling data and compute and 2) theoretical results cannot realistically be transferred to practical scenarios \cite{sutton2019bitter}. 
        We discuss this view in detail in \cref{sec:alt_view}, but emphasize that algorithms with identifiable \glspl{dgp} are used in many scientific and applied domains, including: robotics~\cite{locatello_weakly-supervised_2020,lippe_biscuit_2023}, dynamical systems~\cite{lippe_icitris_2022,rajendran_interventional_2023}, neuroimaging~\cite{himberg2004validating,hyvarinen_unsupervised_2016}, neuroscience \cite{zhou2020learning,schneider2023learnable}, genomics~\citep{morioka_connectivity-contrastive_2023,morioka_causal_2023}, structural biology \cite{klindt2024towards_frontiers}, ant colonies \cite{yao2024unifying}, and climate science \cite{yao_marrying_2024}.
        What is common in \gls{it}-driven applications is the focus on inferring the \textit{correct} (unveiling the underlying science) and \textit{robust} (generalizing, even \gls{ood}) mechanism from the data. Indeed, recent works in \gls{it} showed that \gls{ood} generalization is possible~\citep{brady_provably_2023,brady2025interaction,wiedemer_provable_2023,wiedemer_compositional_2023}. Thus, these practically relevant considerations illustrate the benefit of \gls{it}-driven \gls{ssl} design.

    \begin{table*}[ht]
    \centering
    \setlength{\tabcolsep}{1.6pt}
    \begin{tabular}{lllc} \toprule\midrule
         \textbf{Problem} & \textbf{Domain}& \textbf{Status Quo}
         & \textbf{RQ's}\\\midrule
         \multirow{2}{*}{\textbf{Data augmentations}} & Theory & {\color{orange}?} Augmentations are modeled in the DGP, not in observation space
        & \multirow{2}{*}{ \ref{rq:agnostic_da}}\\
         & Practice & {\color{figgreen}\checkmark} The role of augmentations is understood
         &\\ \midrule
        \multirow{2}{*}{\textbf{Finite data}} & Theory & {\color{figred}\xmark} Finite-sample analysis is almost entirely missing from IT 
        & \multirow{2}{*}{\ref{rq:finite}}\\
        & Practice & {\color{figgreen}\checkmark} Scaling laws show the effect of data set size 
        &  \\ \midrule
        \multirow{2}{*}{\textbf{Data diversity}} & Theory & {\color{figgreen}\checkmark} Data diversity is well-defined in the DGP view of IT 
        & \multirow{2}{*}{\ref{rq:task_diversity}}\\
         & Practice & {\color{figred}\xmark} Data diversity is sometimes entangled with data set size 
         & \\ \midrule
         \multirow{2}{*}{\textbf{Finite time}} & Theory & {\color{orange}?} Crude understanding of (linear) training dynamics, but not tied to IT 
        & \multirow{2}{*}{ \shortstack[2]{\ref{rq:content_style}\\\ref{rq:dyn}}} \\
         & Practice &  {\color{figgreen}\checkmark} Convergence differs across models
         &\\ \midrule
          \multirow{2}{*}{\textbf{Loss saturation}} & Theory &  {\color{orange}?} IT does not consider time, learning dynamics provides insights for linear models
        & \multirow{2}{*}{\ref{rq:dyn}}\\
         & Practice & {\color{figred}\xmark} It is unclear why convergence speeds differ
         & \\ \midrule
         \multirow{2}{*}{\textbf{Inductive biases}}& Theory &   {\color{figred}\xmark} Results missing or not reflecting practical choices
        & \multirow{2}{*}{\ref{rq:suff_nec_ident}}\\
         & Practice & {\color{figgreen}\checkmark} The role of architecture is well-understood
         & \\ \midrule
        \multirow{2}{*}{\textbf{Initialization} }& Theory &  {\color{figred}\xmark} IT does not consider initialization
        & \multirow{2}{*}{\ref{rq:suff_nec_ident}}\\
         & Practice & {\color{figgreen}\checkmark} It is well understood what initialization are useful
         & \\ \midrule
         \multirow{2}{*}{\textbf{Dimensional collapse}} & Theory & {\color{figgreen}\checkmark} The DGP view defines what latents collapse
         & \multirow{2}{*}{\ref{rq:projector}}\\
         & Practice & {\color{orange}?} Computational tricks and regularizers might help collapse
         & \\ \midrule
        \multirow{2}{*}{\textbf{Projector}} & Theory &  {\color{figred}\xmark} No explicit result, some settings reflect a linear projector 
        & \multirow{2}{*}{\ref{rq:projector}}\\
         & Practice & {\color{orange}?} The projector is not eliminated, at most ameliorated
         & \\ \midrule
         \multirow{2}{*}{\textbf{Compositionality}} & Theory & {\color{figgreen}\checkmark} Theoretical results exist for compositional generalization
        & \multirow{2}{*}{\ref{rq:ood}}\\
         & Practice & {\color{figred}\xmark} Empirical methods struggle to generalize compositionally 
         & \\ \midrule
         \multirow{2}{*}{\textbf{CL/non-CL}} & Theory & {\color{figred}\xmark} Theory for non-CL methods is missing
        & \multirow{2}{*}{\shortstack[2]{\ref{rq:ident_non_cl}\\\ref{rq:cl_non_cl}}}\\
         & Practice & {\color{figgreen}\checkmark}Representations are similar
         & \\ \midrule
         \multirow{2}{*}{\textbf{Evaluation}} & Theory &  {\color{figgreen}\checkmark} IT characterizes what latents are learned
        & \multirow{2}{*}{\shortstack[2]{\ref{rq:ssl_eval}\\\ref{rq:ssl_dataset}}}\\
         & Practice & {\color{figred}\xmark} Mostly benchmark-related, principles often missing
         & \\ \midrule
        \bottomrule
    \end{tabular}
    \caption{\textbf{The gaps between\ \gls{ssl} theory and practice:} RQ is a shorthand for research question, {\color{figgreen}\checkmark} denotes a comprehensive understanding; {\color{orange}?} incomplete or related results; and {\color{figred}\xmark} mostly uncharted territory.
    }
    \label{table:blind_spots}
\end{table*}
    
\section{Position: Singular Identifiability Theory as a blueprint to close the gap between SSL theory and practice}\label{sec:pos}
        With the example of SimCLR, we showed how IT can improve our understanding of SSL methods (\cref{subsec:it_simclr}), particularly by making the assumptions on the data explicit via a DGP. However, IT relies on simplifications and cannot explain many practical phenomena, including dimensional collapse~\cite{jing_understanding_2022}, or \gls{ood} behavior.
         \textit{\textbf{Thus, we need to move beyond the current paradigm and address the practical realities of modern machine learning. 
        We call for an extension of IT to what we term Singular Identifiability Theory (SITh), and detail what it can bring to \gls{ssl}.}} Similarly to how Singular Learning Theory (SLT)~\citep{watanabe_algebraic_2009,watanabe_mathematical_2020} was proposed to address, among others, the reality that due to certain symmetries, (the Fisher information matrix of) neural networks can have singularities. We want to emphasize that \gls{sith}, in its current form, is an umbrella term for a \textit{future} theory and a \textit{blueprint}~\cite{bronstein2021geometric} for SSL research. We aim to provide signposts along which the path towards this new theory can be traversed.
        Thus, we review the gaps between SSL theory and practice (\cref{table:blind_spots}). We detail the empirical and/or theoretical evidence in the following subsections, pinpoint potential synergies, and formulate research questions, answering which, we believe, will move the field forward.

    \subsection{The data augmentation gap: unrealistic DGPs}    

         Extensive evaluations demonstrate that data augmentation strategy matters more than the \gls{ssl} method~\citep{morningstar_augmentations_2024}. 
         As we showed with SimCLR in \cref{sec:it}, the \gls{dgp} is a useful construct to model the augmentations---and is at the core of identifiability results in \gls{cl}~\citep{hyvarinen_unsupervised_2016,morioka_independent_2021,zimmermann_contrastive_2021,rusak_infonce_2024}.
         This means that changing an augmentation should be reflected in the DGP, and such changes might make some latents less feasible or impossible to learn~\citep{von_kugelgen_self-supervised_2021,daunhawer_identifiability_2023}. 
         The problem with current IT is that the modeling assumptions in the DGP are \textit{too simplistic}. Using a hypersphere as latent space and simple conditionals such as \gls{vmf} distributions are neither principled nor realistic---for example, they cannot capture common augmentations such as (heavy) crops~\citep{zimmermann_contrastive_2021,rusak_infonce_2024,reizinger_cross-entropy_2024}.

        \begin{researchquestion}\label{rq:agnostic_da}
                \citet{moutakanni_you_2024} suggest that prior knowledge is not required for augmentation design. To what extent is this the case, does this impose performance limitations, and is it possible to develop a theoretical understanding of this phenomenon?
            \end{researchquestion}

    \subsection{The asymptotic data gap: data set and batch size}
            Identifiability results assume infinite data, batch size, and converged, \ie, IT is an \textit{asymptotic} theory---apart from \citet{campi_finite_ident_2002,lyu_finite-sample_2022}.
            In practice, data set size and data or task diversity~\citep{elmoznino_-context_2024,raventos_pretraining_2023} matters\footnote{Though ``diversity" in these contexts often refers to the number of tasks, \eg, in \citet{raventos_pretraining_2023}}. However, it is unclear whether data set size improves performance or it only correlates with properties like diversity in the sense of the ICA literature (intuitively, data coming from diverse environments), which is crucial for identifiablity~\citep{hyvarinen_unsupervised_2016,morioka_independent_2021,rajendran_interventional_2023,khemakhem_variational_2020,reizinger_cross-entropy_2024}.
            More data can \textit{seemingly} improve \gls{ood} performance. However,~\citet{mayilvahanan_does_2024,mayilvahanan_search_2024} showed that the real reason is that the \gls{ood} tasks became in-distribution as they were included in the training data.
            Batch size is also important in practice~\citep{chen_simple_2020,rusak_infonce_2024}, but there are no corresponding theoretical results. Recently, \citet{reizinger_cross-entropy_2024} showed that there is a connection between InfoNCE and DIET, where one difference is calculating the loss over the whole dataset for DIET, whereas InfoNCE uses a mini-batch of data. As both methods are identifiable, this might suggest that the role of batch size can be understood theoretically.

             \begin{researchquestion}\label{rq:finite}
                \citet{lyu_finite-sample_2022} provided a finite-sample analysis on the data diversity condition in nonlinear ICA, whereas \citet{lyu_understanding_2021} analyzed the sample complexity of recovering shared (content) latents in multi-view \gls{ssl}. Can we use such results to provide practically meaningful recommendations on data set size and data diversity to achieve identifiable representations empirically?
            \end{researchquestion}

             \begin{researchquestion}\label{rq:task_diversity}
                Is the number of tasks a model is trained on (called ``task diversity" in \citet{raventos_pretraining_2023}) a causal factor in improving model performance or is it more of a proxy to diversity in the ICA sense~\citep{hyvarinen_nonlinear_2019}, \ie, more tasks presumably mean more diverse tasks? 
                For example, \citet{bansal_smaller_2024} provides evidence that data diversity matters.
            \end{researchquestion}

    \subsection{The finite time gap: learning dynamics and loss saturation}

            Identifiability theory cannot distinguish between the convergence speed of models, as it focuses on models at the global optimum of the loss, \ie, at convergence. 
            As experiments demonstrate, not achieving convergence to the global optimum---\eg, by a saturating loss, when it is very close to the optimum, but small differences can lead to qualitatively different models~\citep{liu_same_2023,reizinger_position_2024}---is often the barrier to high-quality representations~\citep{simon_stepwise_2023,rusak_infonce_2024,von_kugelgen_self-supervised_2021,kadkhodaie2024generalization,roeder_linear_2020}. 
            Another example of this is \textit{grokking}, where the training loss saturates long before the true, generalizing solution is discovered \cite{power_grokking_2022}.
            Recently, theoretical insights started to emerge that provide some guidance on how to improve latent recovery via augmentation design, ensemble models, or hard negative sampling~\citep{eastwood_self-supervised_2023,rusak_infonce_2024}, which might guide further research understanding how such choices affect learning dynamics (potentially in terms of a DGP).
             Current IT aims to recover all latent factors, ignoring the minimality argument of \citet{achille_emergence_2018}, and the practical considerations of efficiency and compression. ``Partial" identifiability results such as those of \citet{von_kugelgen_self-supervised_2021,daunhawer_identifiability_2023} are generally seen in negative light---even though some latent factors might not matter downstream.

             \begin{researchquestion}\label{rq:content_style}
                Why is it the case that sometimes, even though identifiability is only provable for some latent factors, with a large enough latent space and training time, the other latent factors can also be learned (to some extent)~\citep{von_kugelgen_self-supervised_2021}?
             \end{researchquestion}
             
             \begin{researchquestion}\label{rq:dyn}
                 Can we understand what model components, such as vanishing initialization~\citep{simon_stepwise_2023,kunin_get_2024,draganov_importance_2025}, determine the convergence speed differences of SSL methods and how data augmentation design and negative sampling change learning dynamics?
             \end{researchquestion}

    \subsection{The architecture gap: teacher/student networks, initialization, stop gradient}

            The \gls{ssl} model zoo hinges on particular architectural choices like stop gradients and predictor networks, mostly in the non-contrastive domain~\citep{bardes_vicreg_2021,zbontar_barlow_2021,oquab_dinov2_2024}. 
            The necessity of engineering practices and model components needs to be critically evaluated to improve robustness, decrease computational cost, and to make theoretical analysis easier---leading to a potentially wider range of insights. 
             The DIET~\citep{ibrahim_occams_2024} instance discrimination method was proposed as a simplified \gls{ssl} pipeline to avoid the theoretically not well-understood components of \gls{ssl} (\eg, stop gradients). DIET performs comparably to other \gls{ssl} methods and is proven to identify a cluster-centric DGP~\citep{juhos2024diet,reizinger_cross-entropy_2024}.
             These works demonstrated that \gls{ssl} can provably work without many computational tricks, hopefully helping practitioners to design more efficient pipelines in the future.

             \begin{researchquestion}\label{rq:suff_nec_ident}
                \citet{morningstar_augmentations_2024} showed that many different algorithmic knobs do not substantially alter the downstream performance of self-supervised representations. How can we use this observation to develop theoretical results showing
                what are the necessary and sufficient architectural and data components to prove identifiability?
            \end{researchquestion}

           \begin{figure*}[t]
        \centering
        \includesvg[width=0.8\textwidth,keepaspectratio]{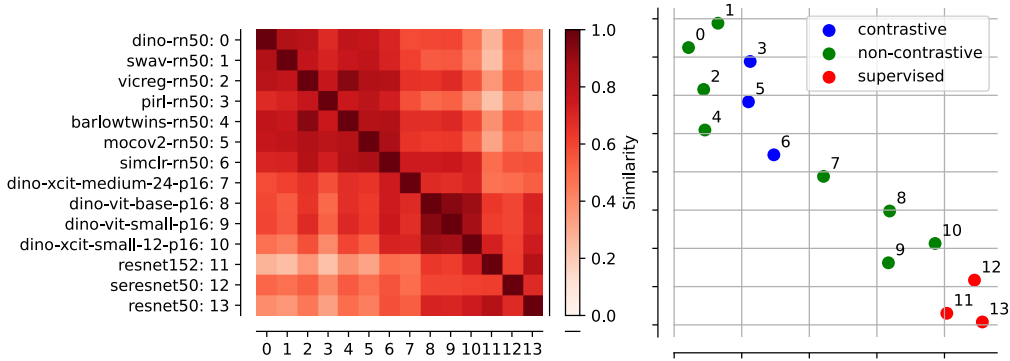}
        \caption{\textbf{Representational similarities between contrastive and non-contrastive methods trained on ImageNet-1k:} 
        \textbf{(Left):} the similarities are calculated with an RBF-kernel--based CKA with $\sigma=0.4$. We include supervised baselines for reference.
        Reproduced from \citep{ciernik_training_2024} with the authors' permission.
        \textbf{(Right):} a two-dimensional UMAP embedding of similarity matrix colored by method types (contrastive, non-contrastive and supervised).
        }
        \label{fig:similarity_matrix_cka_kernel_rbf}
    \end{figure*}

          \subsection{The projector gap: dimensional collapse}
            The projector phenomenon refers to better latent recovery/downstream performance at the other-than-the-ultimate layers of the trained network~\cite{chen_simple_2020}. That is, some dimensions are collapsed \citep{jing_understanding_2022}. Note that this is distinct from \textit{neural collapse}~\citep{papyan2020prevalence} where the model learns one-hot predictions of the training data rather than learning the correct conditional probabilities as assumed, e.g., in \citet{reizinger_cross-entropy_2024}. The projector is seen as ``wasting" resources by training the last (few) layers, and its mitigation is actively researched~\citep{tian_understanding_2022,bordes_guillotine_2023,jing_understanding_2022,song_towards_2023,bizeul_probabilistic_2024,xue_investigating_2024,saunshi_mathematical_2021}.
            The projector is an accidental phenomenon---\ie, no one designs an \gls{ssl} pipeline by saying ``the last $k$ layers are the projector, \ie, I expect them to be useless for downstream performance."
            What is a deliberate choice is to use a linear or a nonlinear projector. 
            Interestingly, a nonlinear projector is not necessarily better---see comparison in, \eg, \citep[Fig.~2]{mialon_variance_2022}.
            We argue that there is a possible theoretical explanation of both 1) why the projector phenomenon occurs, and 2) why a linear projector is often better than a nonlinear one.
             \citet{roeder_linear_2020,hyvarinen_unsupervised_2016,reizinger_cross-entropy_2024} assumed that the trained neural network is a composition of a nonlinear encoder $\mathbf{f}$ and a linear projection matrix $\mathbf{W}, $ yielding $\mathbf{W}\circ\mathbf{f}.$ Then they prove identifiability of $\mathbf{z} = \mathbf{f}(\mathbf{x})$ in a log-linear model, where the $\mathbf{z}$ is fed through a softmax to yield the loss. 
             We hypothesize that, in certain cases, this log-linear model can describe and explain the (linear) projector's occurrence. So far, this potential link between identifiability results and the projector phenomenon has eluded the community, and requires further research.

            \begin{researchquestion}\label{rq:projector}
                \citet{roeder_linear_2020,reizinger_cross-entropy_2024} show that under a specific \gls{dgp}, the key construct for identifiability is a log-linear model, \ie, estimating cross entropy with a softmax. Does this provably explain the (linear) projector phenomenon? Is it possible to develop an identifiability result without the projector?
            \end{researchquestion}

    \subsection{The generalization gap: OOD and compositionality}
        Identifiability results have recently been extended to specific \gls{ood} scenarios in computer vision, based on compositionality~\citep{brady_provably_2023,brady2025interaction,wiedemer_compositional_2023,wiedemer_provable_2023,lachapelle_additive_2023}. These results exploit that composition of attributes and objects is often straightforward to model in a DGP. Their proofs rely on imposing structural constraints such as additivity on the generative model. 
        Most empirical representation learning methods do not generalize compositionally~\citep{montero_role_2021,montero_lost_2024,schott_visual_2021,wiedemer_provable_2023}, emphasizing the need for theoretical insights to design better SSL methods. Interestingly, in the language domain, both the first theoretical results~\citep{ahuja_provable_2024,xie_explanation_2022}, and empirical techniques and observations~\citep{bansal_end--end_2022,ruoss_randomized_2023,meszaros_rule_2024} exist showing that OOD generalization might be possible, at least in terms of length generalization or rule extrapolation.

        \begin{researchquestion}\label{rq:ood}
            Is it possible to provide OOD or compositional generalization results across data modalities, such as in image-text models? If yes, is there a qualitative or quantitative difference \wrt the inductive bias of each modality?
        \end{researchquestion}
            
     \subsection{The Platonic representation gap: the contrastive and non-contrastive dichotomy}\label{subsec:pos_unify}

        A prevalent categorization of \gls{ssl} is the contrastive/non-contrastive dichotomy. While categorizations like these are undoubtedly valuable, they might obscure the overall trends within the \gls{ssl} model landscape. However, recent initiatives are striving to bring unity to the field of \gls{ssl}~\citep{balestriero_cookbook_2023,morningstar_augmentations_2024,cabannes_ssl_2023}.
         \citet{garrido_duality_2022} showed that both method families contrast some properties; for this reason, they propose the names \textit{sample-contrastive} (for \gls{cl}) and \textit{dimension-contrastive} (for non-contrastive).
        \citet{balestriero_contrastive_2022} analyzed contrastive and non-contrastive paradigms with a spectral embedding approach, showing that such methods correspond to specific local and global spectral embedding methods (e.g., SimCLR was shown to be equivalent with ISOMAP~\cite{tenenbaum2000global} under specific assumptions). Similarly, t-SNE can be made contrastive with minimal modifications \cite{bohm2022unsupervised}.
        There is also evidence that both paradigms relate to entropy and mutual information estimation~\citep{liu_self-supervised_2022}.
        \citet{wang_understanding_2020} interpreted InfoNCE as an interaction between uniformity (\ie, entropy estimation), and alignment (\ie, invariance \wrt augmentations).
        The InfoNCE family is shown to minimize cross-entropy~\citep{zimmermann_contrastive_2021,rusak_infonce_2024}; the same holds for the (non-contrastive) instance-discrimination method called DIET, which is also related to InfoNCE~\citep{ibrahim_occams_2024,reizinger_cross-entropy_2024}. 
         These principles are also present in non-contrastive methods, though usually via regularizers. For example, 
        VICReg~\citep{bardes_vicreg_2021} also estimates a lower bound on entropy~\citep{shwartz-ziv_what_2022}. 
        One distinction is the absence of an explicit \gls{dgp} in most non-CL methods. That is, the DGP is mostly defined implicitly, with the exception of~\citep{reizinger_cross-entropy_2024}.

        \begin{tcolorbox}[colback=figred!15!white,colframe=figred!75!black, title={The contrastive and non-contrastive split posits a sterile dichotomy }]
            Contrastive and non-contrastive methods are driven by the same principles, thus, we need to consider their similarities, and not focus on their differences as they represent different means to the same end.
        \end{tcolorbox}

            Experimental evidence demonstrates that (identifiable) contrastive and non-contrastive representations fare similarly on some downstream task~\citep{morningstar_augmentations_2024} and that their representations can be highly similar~\citep{ciernik_training_2024}---\cf \Cref{fig:similarity_matrix_cka_kernel_rbf} and also note similarity is not universal, \eg, ViT-based models are less similar. However, identifiability guarantees for non-contrastive methods are mostly missing---with the exception of DIET~\citep{ibrahim_occams_2024,reizinger_cross-entropy_2024}. The empirical similarity suggests that some non-contrastive methods might be identifiable.

            \begin{researchquestion}\label{rq:ident_non_cl}
                Based on the similarity of contrastive and non-contrastive representations~\citep{morningstar_augmentations_2024,ciernik_training_2024} and the identifiability of contrastive methods, can we prove identifiability for non-contrastive methods?
            \end{researchquestion}
        
           \begin{researchquestion}\label{rq:cl_non_cl}
                The representations of contrastive and non-contrastive methods are often similar~\citep{ciernik_training_2024} (\cf \cref{fig:similarity_matrix_cka_kernel_rbf}), though their convergence speed is highly distinct~\citep{simon_stepwise_2023}. Can we theoretically understand in which (data, model size, \etc) regimes which method is expected to perform best?
            \end{researchquestion}

       \subsection{The evaluation gap: going beyond ImageNet}

            Most \gls{ssl} methods evaluate performance on ImageNet~\citep{krizhevsky2012imagenet} downstream classification, without acknowledging that such an evaluation can only falsify claims about classification-related latents. For example, if orientation is not required for classification, then a more universal representation (\eg, one capturing orientation) will not perform better---though this makes a difference in terms of ``universality" of the representation. Indeed, SSL methods capturing more latents sometimes have (slightly) lower downstream classification accuracy~\citep{rusak_infonce_2024}, which might suggest their infeasibility---though that depends on what they will be used for.
            \textit{We need to question the practice of measuring the ``universality" of representation with a single scalar on a highly specialized task.} 
            A good example of a more comprehensive evaluation is DINOv2~\citep{oquab_dinov2_2024}.
            Another gap, especially for evaluating identifiability claims, is the lack of large-scale data sets with latent information---proxies such as ImageNet-X~\citep{idrissi2022imagenetx} exist, though they only have binary labels.
            Recent works started using aggregate statistics (\eg, rank conditions) to evaluate the representation quality~\citep{agrawal_alpha-req_2022,garrido2023rankmeassessingdownstreamperformance,thilak2023lidarsensinglinearprobing}. These methods seem to be somewhat predictive of even \gls{ood} performance, we lack a theoretical understanding when and why this is the case.

        \begin{tcolorbox}[colback=figred!15!white,colframe=figred!75!black, title={Evaluating self-supervised representations requires more than ImageNet classification}]
             \gls{sith} can improve SSL evaluations by using a DGP to focus on the relevant latent factors and design more principled benchmarks.
        \end{tcolorbox}

         \begin{researchquestion}\label{rq:ssl_eval}
                How can we evaluate the ``universality" of self-supervised representations for robustness? \Ie, how should we move away from downstream classification on ImageNet? What are the \textit{set} of downstream tasks we need to evaluate SSL methods?
            \end{researchquestion}

        \begin{researchquestion}\label{rq:ssl_dataset}
             How can we develop ImageNet-scale (synthetic) datasets with latent information to evaluate the identifiability of self-supervised representations? Can we use rendering pipelines akin to DisLib~\citep{locatello_challenging_2019} to develop more principled benchmarks?
        \end{researchquestion}

    \subsection{Position summary: We need an empirically grounded identifiability theory for \gls{ssl}}\label{subsec:pos_summary}
        To improve our understanding of the (implicit) operating principles of SSL methods such as SimCLR~\citep{zimmermann_contrastive_2021} (\cref{subsec:it_simclr}). Recent IT results also improved our understanding of compositional generalization~\citep{brady_provably_2023,brady2025interaction,wiedemer_compositional_2023,wiedemer_provable_2023,lachapelle_additive_2023}.
        Yet, current IT cannot fully explain empirical observations in SSL. To quote Nobel-prize--winning physicist Richard P. Feynman: 
            \textit{“It doesn't matter how beautiful your theory is, if it doesn't agree with the experiment, it's wrong.”
            }
         IT models the data via an underlying \acrfull{dgp}, however, the design of the DGP does not exploit guidance from empirical observations, which leads to 
        presumably the biggest critique of identifiability theory:
            \textit{Identifiability guarantees are not valuable per se, only to the extent they can describe empirical phenomena, or---and only time will tell this---they aid the development of such theories that explain such phenomena.}

        When a theorist constructs a \gls{dgp}, the focus should not only be on identifiability but also on the match with reality \cite{klindt_towards_2021}.
        Vice versa, when a practitioner sets up a pre-text task for \gls{ssl}, they should think about the implicit \gls{dgp} that they might be assuming and turn to theory to ask whether this setup results in an identifiable \gls{dgp} (and estimation procedure).
        Constructing a \gls{dgp} on the hypersphere is common in theory~\cite{zimmermann_contrastive_2021,reizinger_cross-entropy_2024}. However, it is unclear how well this actually corresponds to a realistic \gls{dgp} (for images).
        Analogously, building a pre-text around data augmentations is an empirically successful construct, however, it will only result in theoretic identifiability if the data augmentations cover all degrees of freedom in the data---a rather unrealistic assumption, mirrored by findings on the importance of the right augmentations~\citep{chen_simple_2020,oquab_dinov2_2024,morningstar_augmentations_2024}.

\section{Alternative Views}\label{sec:alt_view}

    Recent years' breakthroughs in AI were often fueled by scaling up models, compute, and data \cite{sutton2019bitter}. Though empirical algorithmic improvements are also crucial for breakthroughs like the recent DeepSeek-R1~\cite{deepseekai2025deepseekr1incentivizingreasoningcapability} reasoning model. The scaling-related observations were documented in so-called scaling laws~\citep{hernandez_scaling_2021,zhai_scaling_2022,henighan_scaling_2020,hestness_deep_2017,kaplan_scaling_2020}, analyzing trends in loss reduction. 
    Especially in the case of \glspl{llm}, scaling data and compute is still significant for progress, maintaining the popularity of this opinion in the community in general.
    
    \textbf{Counterarguments in SSL:} Realistically, engineering efforts are crucial to solve the problems of real-world \gls{ssl} systems. However, we are seeing the \gls{ssl} community shift beyond scaling.\footnote{Personal observations at the \hyperlink{https://sslneurips2024.github.io/}{NeurIPS 2024 Workshop: Self-Supervised Learning - Theory and Practice}}
    After scraping the majority of data from the Internet, and pushing the boundaries of compute beyond previously unimaginable limits, we need a new approach.  \citet{sorscher_beyond_2022} theoretically showed that scaling laws can sometimes be overcome via pruning. More data can also imply that \gls{ood} tasks become in-distribution~\citep{mayilvahanan_does_2024,mayilvahanan_search_2024}, qualitatively changing what we as a community thought about the true role of scaling data.
    We propose an identifiability theoretical framework as an alternative to the scaling view, still prevalent in \gls{llm} research. 
    We believe that having a more principled approach towards \gls{ssl}, which incorporates the empirical observations, is the way to move \gls{ssl} forward.
    Particularly, improving \gls{ood} generalization performance in \gls{ssl} is hard to do empirically. First, most empirical representation learning methods, even despite aiming to learn disentangled representations, do not generalize compositionally~\citep{montero_role_2021,montero_lost_2024,schott_visual_2021,wiedemer_provable_2023}. Second, it is hard to get the intuition without using a model like a \gls{dgp}.
    We believe that \acrfull{sith} can help overcome these limitations, and lead to more efficient, principled, and robust \gls{ssl} methods. 

\section{Conclusion}
    Our position paper hit a critical note to highlight what we conceive as a big barrier before closing the gap between SSL theory and practice. We argued that thinking in terms of an empirically-grounded \acrfull{dgp} can accelerate SSL research. We formulated our suggestion in terms of an extension to \acrlong{it}, called \acrfull{sith}, to foster synergies between the two subfields. To provide a common starting point, we included an intuitive introduction of identifiability for practitioners (\cref{sec:it}), demonstrating how SSL practice can benefit from IT.
    However, more research needs to be done, illustrated by the many open questions in both theory and practice (\cref{table:blind_spots}).
    As we are close to hitting the limits of scaling-driven advances, we need to turn towards more principled, theoretically grounded approaches. This is especially important when machine learning models are deployed in safety-critical environments such as autonomous driving or healthcare. We believe that \gls{sith} facilitates asking better research questions, many of which we shared, to drive SSL forward
    by improving how we design, evaluate, and understand SSL algorithms.

%% file: appendix.tex
\section{On absolute vs relative identifiability}\label{sec:app_abd_rel_ident}

    We need to tackle one nuance when talking about identifiability and the alignment of representations: namely, what do we mean by identifiability? There are two notions in the literature, which we will call \textit{relative} and \textit{absolute identifiability}. 
    These are defined as follows: 

    \begin{definition}[Absolute identifiability of the ground-truth representation]\label{def:abs_ident}
        Given a ground-truth probabilistic model $p_\theta,$ and a learned family of probabilistic models parametrized by a neural network at the optimum of its training loss $p_{\hat{\theta}} : \hat{\theta} \in \hat{\Theta},$ the inferred model identifies the ground-truth representation if $\forall  \hat{\theta} \in \hat{\Theta}$ and a particular equivalence relationship $\sim_{abs}$ it holds that $p_{\hat{\theta}} \sim_{abs} p_{{\theta}}$.
    \end{definition}

    That is, absolute identifiability means that there is an equivalence relationship (\eg, permutation or scaling) between the learned (family of) representations and the ground-truth one. ``Family" refers to, \eg, training the same model with different seeds. This identifiability notion is more prevalent in the \gls{ica} literature~\citep{hyvarinen_nonlinear_1999,hyvarinen_independent_2013,hyvarinen_identifiability_2023}.

    On the other hand, relative identifiability posits an equivalence relationship between two trained models, \ie, it does not make any claims about the ground-truth representation. 

    \begin{definition}[Relative identifiability of a pair of learned representations]\label{def:rel_ident}
        Given a pair of learned probabilistic model families $p_{\theta_1}, p_{\theta_2}: \theta_1\in \Theta_1, \theta_2\in \Theta_2$ parametrized by (not necessarily the same) neural networks at the optimum of their respective training losses, they are said to be identified in a relative sense if there exists an equivalence relationship $\sim_{rel}$ such that $\forall \theta_1\in \Theta_1, \theta_2\in \Theta_2: p_{\theta_1} \sim_{rel} p_{\theta_2}$.
    \end{definition}

\section{The Platonic Representation Hypothesis revisited}\label{sec:prh}

    \citet{huh_platonic_2024} argued that the 
        i) function class,
        ii) dataset, and
        iii) regularization,
    determine the convergence of representations, which are important in \gls{ssl}~\citep{cabannes_ssl_2023,balestriero_cookbook_2023,morningstar_augmentations_2024}.
    By synthesizing evidence from the identifiability literature, we showcase the mathematical results answering some of these questions. However, we will also show the negative results and the limits of current IT, enabling us to precisely specify future research directions. Before that, we emphasize that the \gls{prh}'s claims are about the similarity of two \textit{learned} representations (such as in \citet{roeder_linear_2020}), whereas most identifiability results are about an \textit{absolute} sense between a learned model and the assumed underlying \gls{dgp}---we discuss the differences in \cref{sec:app_abd_rel_ident}.   

    \paragraph{Function class and model capacity.}
        Restricting the function class is a conventional technique to prove identifiability~\citep{hyvarinen_nonlinear_1999,locatello_challenging_2019,bloem-reddy_probabilistic_2020,buchholz_function_2022}. That is, in some cases no restriction can imply non-identifiability, indicated by the impossibility results of~\citet{locatello_challenging_2019,hyvarinen_nonlinear_1999}. A restriction of the functions class in form of a log-linear model was also shown to be key for recent identifiability results in (self-)supervised learning~\citep{hyvarinen_unsupervised_2016,hyvarinen_nonlinear_2019,roeder_linear_2020,reizinger_cross-entropy_2024,shimizu_linear_2006,hoyer_nonlinear_2008,lachapelle_gradient-based_2020,gresele_independent_2021}. Compositional generalizationalso hinges on particular (mainly additive) decoder structures~\citep{lachapelle_additive_2023,brady_provably_2023,wiedemer_provable_2023,wiedemer_compositional_2023,brady2025interaction,mahajan_compositional_2024,ahuja_interventional_2022}
        Such results indicate that the \gls{prh}'s validity is potentially restricted to a specific model class. The investigation of \citet{ciernik_training_2024} also suggests a varying degree of similarity even between self-supervised models (\cref{fig:similarity_matrix_cka_kernel_rbf}).

    \paragraph{Data set and training objective.}
        Many self-supervised models optimize an objective that corresponds to the cross entropy between the distributions of the underlying and the learned \glspl{dgp}~\citep{zimmermann_contrastive_2021,huh_platonic_2024,reizinger_cross-entropy_2024}.
        This implies that, \eg, changing the way positive pairs are generated implies adapting the loss function to avoid a mismatch~\citep{zimmermann_contrastive_2021,rusak_infonce_2024}.
        \citet{von_kugelgen_self-supervised_2021} proved that when some latent factors are not changed by the augmentations, then only those are provably identifiable---others collapse, as observed in the projector phenomenon~\citep{chen_simple_2020,rusak_infonce_2024,jing_understanding_2022}. This supports empirical observations about the importance of augmentations~\citep{morningstar_augmentations_2024}, and suggests that different sets of augmentations can have different ``Platonic ideals".

    \paragraph{Regularization.}
        \citet{huh_platonic_2024} posited that deep networks are biased towards learning simple functions~\citep{simplicitybiasSGD}. 
        In a discussion about the non-identifiability of \glspl{llm}, \citet{reizinger_position_2024} noted that the theoretical understanding of a wide range of inductive biases is still missing---this applies to \gls{ssl} in general, too---at least for inductive biases beyond the ones expressed via the loss function, the function class, or data augmentations. \citet{rusak_infonce_2024} suggested an anisotropic conditional can worsen the learning dynamics of high-variance factors, though more work is required to precisely understand nonlinear learning dynamics in \gls{ssl}~\citep{simon_stepwise_2023}.

    \paragraph{Summary.}
        In our view, the \gls{prh} made a first important step formalizing the similarities of learned representations. However, it does not provide a clear theoretical framework to precisely investigate such questions